\pdfoutput=1

\documentclass[11pt]{article}

\usepackage{acl}

\usepackage{times}
\usepackage{latexsym}

\usepackage[T1]{fontenc}

\usepackage[utf8]{inputenc}

\usepackage{microtype}

\usepackage{inconsolata}

\usepackage{amsmath}
\usepackage{comment}
\usepackage{graphicx}
\usepackage{tablefootnote}
\usepackage{cprotect}
\usepackage{multirow}
\usepackage{hyperref}
\usepackage{soul}

\usepackage{tablefootnote}

\title{Simul-LLM: A Framework for Exploring High-Quality Simultaneous \\ Translation with Large Language Models}

\author{Victor Agostinelli \: \: Max Wild \: \: Matthew Raffel \\ 
{\bf Kazi Ahmed Asif Fuad} \: \: {\bf Lizhong Chen} \\
Oregon State University \\
\texttt{\{agostinv, wildma, raffelm, fuadk, chenliz\}@oregonstate.edu}
}

\begin{document}
\maketitle

\begin{abstract}
Large language models (LLMs) with billions of parameters and pretrained on massive amounts of data are now capable of near or better than state-of-the-art performance in a variety of downstream natural language processing tasks. Neural machine translation (NMT) is one such task that LLMs have been applied to with great success. However, little research has focused on applying LLMs to the more difficult subset of NMT called simultaneous translation (SimulMT), where translation begins before the entire source context is available to the model. In this paper, we address key challenges facing LLMs fine-tuned for SimulMT, validate classical SimulMT concepts and practices in the context of LLMs, explore adapting LLMs that are fine-tuned for NMT to the task of SimulMT, and introduce \textit{Simul-LLM}\footnotemark, the first open-source fine-tuning and evaluation pipeline development framework for LLMs focused on SimulMT. 
\end{abstract}

\footnotetext{\url{https://github.com/OSU-STARLAB/Simul-LLM}}

\section{Introduction}
Modern large language models (LLMs) contain at least several billion and up to trillions of parameters and are remarkably capable across a wide range of tasks. Pretrained on humongous amounts of unlabeled data, they have demonstrated incredible emergent capabilities. With minor prompt adjustments, such as including instructions and examples, LLMs are often capable of near state-of-the-art performance set by highly customized solutions. The performance of these models is further enhanced when fine-tuned for specialized downstream tasks, sometimes exceeding the performance of previously cutting-edge solutions. Given their rapidly evolving capabilities, LLMs and their application have become a focused topic of research within NLP academia \citep{zhao2023survey}.

One popular downstream task for LLMs is text-to-text neural machine translation (NMT), which focuses on taking an input sequence in a given language and outputting a translation in another language. Typically, the entire source context is available at the start of translation for NMT. A particularly challenging subset of NMT is known as simultaneous translation (SimulMT), where the model begins translation without having access to the entire source sequence, and the translation progresses as the remaining source sequence is incrementally provided. For languages that are syntactically and structurally similar, near-NMT performance is fairly achievable, but for language pairs that differ significantly in structure, traditional models struggle to balance high-quality translations with delay for additional source context. This balance is typically achieved via a fixed or adaptive read-write schedule, with one of the most popular and longstanding fixed schedules being the \textit{wait-k} policy \citep{stacl2019}, where the target translation hypothesis lags behind the incrementally available source sequence by \textit{k} words or subwords. 



While LLMs have been applied to and studied actively in NMT, their application to simultaneous translation has been lagging. This is in part due to a few challenges LLMs face when applied to SimulMT that are non-trivial to address. First and foremost, it is unclear how well LLMs, which are pretrained and usually fine-tuned under the assumption that the prompt is completely provided and static before generation, will adapt to an application space where the prompt dynamically changes as the simultaneous scheduler elects to read from the source sequence. Second, multiple approaches exist to enable LLMs for SimulMT and it is challenging to intuit which approach will perform best. For example, one could adapt LLMs fine-tuned for NMT (hereafter referred to as \textit{NMT LLMs}) to SimulMT during inference, although how well such models will deal with the source context availability mismatch between fine-tuning (full sentence) and inference (partial sentence) is nebulous. Alternatively, one could fine-tune LLMs directly for SimulMT (hereafter referred to as \textit{SimulMT LLMs}), but new prompt structuring is likely needed to match inference SimulMT behavior during fine-tuning exactly. Finally, it is also unclear how well previously understood concepts in existing SimulMT work, such as higher fine-tuning \textit{wait-k} values increasing generalizability, will apply to SimulMT LLMs.


This paper seeks to address the above problems and contributes to the process of applying LLMs to SimulMT in the following major ways: 

\begin{itemize}
    \item We develop Simul-LLM, the first open-source fine-tuning and evaluation pipeline development framework for SimulMT LLMs, which seamlessly wraps around and interfaces with popular libraries for LLMs and SimulMT. This framework serves as a foundation for research on SimulMT LLMs that the community can employ and extend for a wide range of future work on LLM-based simultaneous translation.
    \item With the aforementioned framework, we explore the feasibility of adapting LLMs fine-tuned for NMT to SimulMT under a few decoding strategies and the classical \textit{wait-k} fixed translation scheduler. Generally, we find that NMT LLMs demonstrate good performance during SimulMT inference which can be somewhat boosted by more complex decoding strategies.
    \item We propose an alternative prompt structuring approach to commonly employed NMT prompts that bridges the gap between the fine-tuning and inference environment, assuming a \textit{wait-k} schedule, and we validate this via the Simul-LLM framework. We elaborate on counter-intuitive results that we observe and provide a base of exploration for future research to employ. Along these lines, we also validate that higher \textit{wait-k} values employed during SimulMT fine-tuning do increase \textit{wait-k} generalizability and boost translation quality across the board during SimulMT inference.
\end{itemize}


\section{Background and Motivation}
\label{sec:background}
While a range of work is relevant to this paper, we will only provide a focused and high level review of large language models applied towards machine translation and simultaneous translation as an application space. Readers interested in additional details should engage further with cited works in these areas.

\subsection{Large Language Models for Neural Machine Translation}
\label{sec:llms}
LLMs are capable of effectively zero-shot sentence-to-sentence neural machine translation (NMT) \citep{vilar-etal-2023-prompting}, but their performance can still be improved via simple techniques. Prompt construction has been demonstrated to be critical to LLM performance, both before and after fine-tuning \cite{zhang2023prompting}. One-shot or few-shot performance via In-Context Learning (ICL) can produce near competitive results with fine-tuned LLMs for translation and can even be employed to enhance fine-tuned model performance \citep{vilar-etal-2023-prompting,xu2023paradigm}. 

One particularly interesting area of study related to LLMs applied towards NMT remains whether or not to fully fine-tune a given model or engage in Parameter-Efficient Fine-Tuning (PEFT) \citep{peft}. Early work in this area demonstrated some potential for smaller models \citep{ustun-cooper-stickland-2022-parameter}, and the accessibility that PEFT provides designers in terms of fine-tuning on low-to-mid performance hardware setups renders it desirable. One of the most popular forms of PEFT freezes a LLM's weights and adds Low-Rank Adaptation (LoRA) \citep{hu2022lora} adapters between layers\footnote{Other forms of PEFT exist, but adapter-based PEFT is extremely common so we refer to adapter-based PEFT simply as PEFT hereafter.}. While fully fine-tuned NMT LLMs tend to suffer from some level of catastrophic-forgetting \citep{Kirkpatrick_2017}, intuitively, PEFT-based NMT LLMs should not suffer from any loss of off-task performance, as adapters can be loaded or detached depending on whether or not a given user is prompting for a translation. Given these factors, PEFT is an attractive option for NMT LLMs.

\subsection{Simultaneous Translation}
\label{sec:simulmt}
As a subset of typical NMT, simultaneous translation (SimulMT) focuses on engaging in translation (write decisions) while balancing the amount of available source context (read decisions) to reduce translation lagging behavior/latency. This necessarily increases the difficulty of translating a sequence in one language to a sequence in another language, especially when structural and/or syntactical differences exist in the language pair. As a brief example, we can consider translating from a subject-verb-object (SVO) language, such as English, to a subject-object-verb (SOV) language like German. Based on the available source context in English, we may have to guess at the translation in German without access to the necessary context to effectively make that prediction because of the aforementioned syntactical differences in the language pair.


There are two classical, high level approaches to scheduling the write and read decisions of SimulMT, those being static schedules like \textit{wait-k} \citep{stacl2019} or adaptive schedules which are flexible and learned, such as variants of monotonic multi-head attention, adaptive \textit{wait-k} compositions,  \textit{wait-if-worse}, decision state assisted SMT, and others \citep{grissom-ii-etal-2014-dont, gu2017learning, mma2019, zheng2020simultaneous}. \textit{Wait-k} remains a particularly popular baseline strategy given its ease of application during training and during inference. It functions by retaining a \textit{k}-lagging factor between the source context (either in tokens or in words) $\textbf{x}$ and the translation hypothesis $\textbf{y}$. We can model a typical \textit{wait-k} schedule's probability of generating a given output sequence, provided some source sequence, with Equation \ref{eq:waitk}:

\begin{equation}
p(\textbf{y}, \textbf{x}) = \prod^{|\textbf{y}|}_{i=1}p( \textbf{y}_i|\textbf{y}_{<i}, \textbf{x}_{<min(i+k, |\textbf{x}|)})
\label{eq:waitk}
\end{equation}

Under circumstances or in environments where additional computational latency is acceptable, variations of beam search have been applied in simultaneous scenarios. Speculative Beam Search (SBS) \citep{zheng-etal-2019-speculative} is one such example where, at a high level, each translation step attempts to speculatively translate future steps for some number of beams, eventually selecting a single token or word (the first one) from the most likely beam for some beam length. When applied to \textit{wait-k}, single token or word SBS can be modeled via Equation \ref{eq:waitk_SBS}, where $w$ is the length of the beam and $\Hat{\textbf{y}}_{i+1:w}$ represents the speculative beam of maximum joint probability:

\begin{equation}
    p(\textbf{y}, \textbf{x}) = \prod^{|\textbf{y}|}_{i=1}p( \textbf{y}_i|\Hat{\textbf{y}}_{i+1:w}, \textbf{y}_{<i}, \textbf{x}_{<min(i+k, |\textbf{x}|)})
\label{eq:waitk_SBS}
\end{equation}

Chunk-wise variations are also possible, where for multiple consecutive translation steps, or write decisions, words or subwords from the last determined most-likely beam are employed to cut down on computational latency.

\subsection{Motivation for Applying LLMs to SimulMT}
LLMs are clearly capable of state-of-the-art NMT via a variety of techniques. Extending them towards SimulMT is a natural step, mirroring the progression of classical SimulMT research (i.e. seeking SimulMT improvements from NMT novelties). In this case, smaller language models (LM) have been employed to augment SimulMT efforts before, although usually this augmentation is limited to rendering the translation scheduler more discerning \citep{indurthi-etal-2022-language}. Instead of just augmenting translation schedulers with smaller, bi-directional LMs, we explore using generative LLMs as direct SimulMT models in this work. 

Intuitively, this results in the LLM providing a deeper understanding of source and target languages (via their pretraining on massive, somewhat multilingual corpora) than is typical for most SimulMT translation models. This can matter when encountering context alignment obstacles, where necessary source context is missing for a given translation time-step. In such situations, SimulMT models must infer based on what context does exist and past experiences to determine what the most likely next translated word will be, a task LLMs are naturally suited for.

\section{Simul-LLM: an Open-Source SimulMT LLM Fine-tuning Framework}
SimulMT is an underexplored application space for LLMs, at the moment, and there is plenty of room to improve performance. To facilitate the rapid development of solutions for SimulMT LLMs (fine-tuned for SimulMT) or NMT LLMs (fine-tuned for NMT) adapted for SimulMT, we choose to develop and provide an open-source framework written in PyTorch for researchers to actively employ for future experiments. We call this framework \textit{Simul-LLM}, and it bridges the gap between the development of fine-tuning agents via popular libraries and proper SimulMT evaluation. Simul-LLM is poised to support both selective classical SimulMT systems as well as fine-tuning and evaluation for a variety of LLM systems such as Falcon \citep{falcon40b}, LLaMa \citep{touvron2023llama} and Mistral-based \citep{jiang2023mistral} models. The high-level components of Simul-LLM include a fine-tuning wrapper and a SimulMT evaluation agent for every supported LLM; in the case of classical models, only the evaluation agent is supported in-framework.

\subsection{Fine-tuning Wrapper and Features}
\begin{figure}[t]
    \centering
    \includegraphics[scale=0.27]{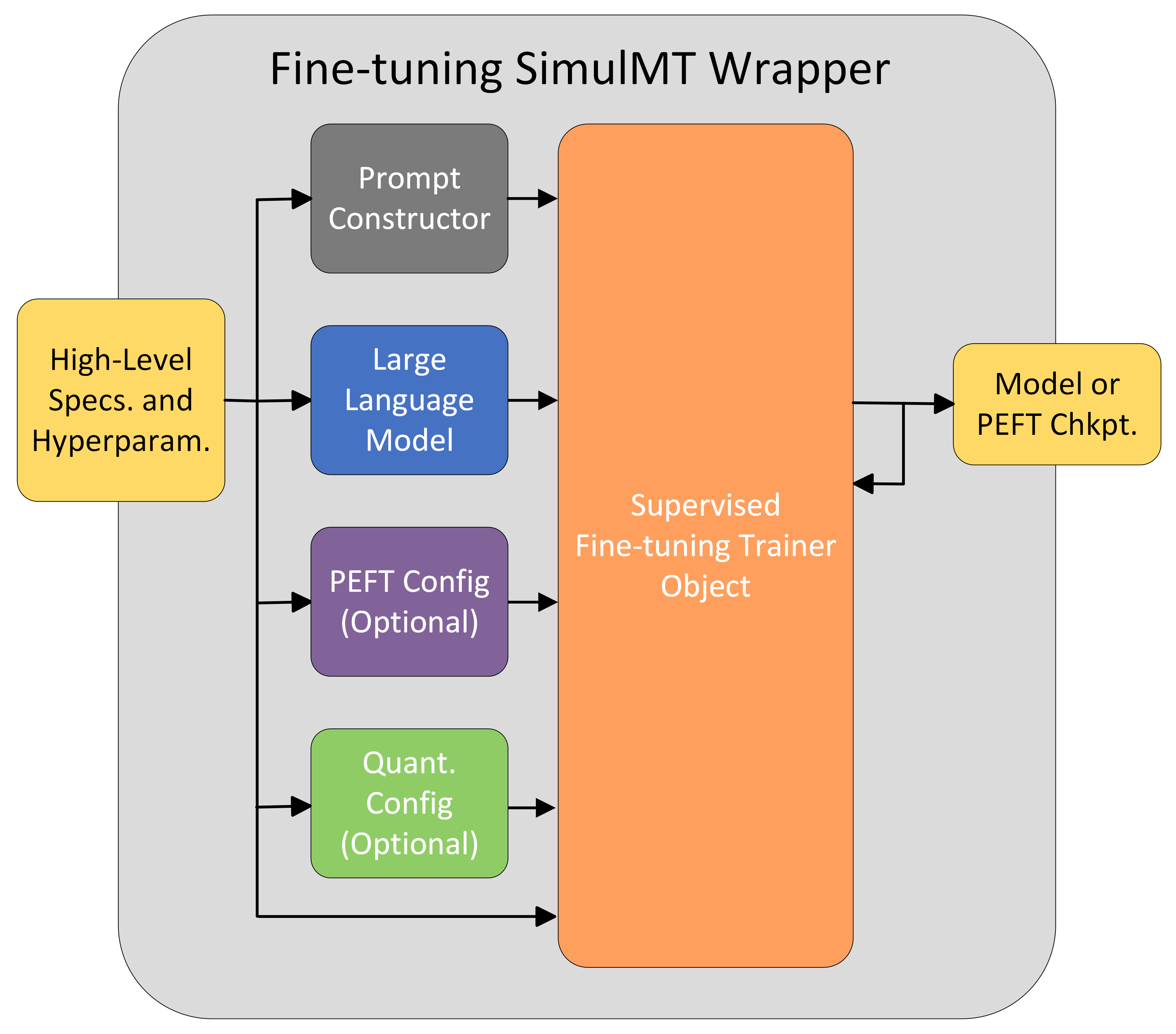}
    \cprotect\caption{Depiction of the Simul-LLM fine-tuning wrapper framework. High level specifications and hyperparameters are passed to the wrapper on instantiation, which employs a specified prompt constructor, instantiates a specified LLM foundational model, optionally constructs a PEFT config, and optionally constructs a quantization config via \verb|BitsAndBytes|.}
    \label{fig:finetune}
    \vspace{-1em}
\end{figure}

The fine-tuning wrapper of Simul-LLM is constructed with a focus on simplicity and extensibility. It is depicted in Figure \ref{fig:finetune} and supports the following set of user-friendly features.

\medskip
\noindent\textbf{LLM Support and Extensibility}: 
The proposed Simul-LLM currently supports Falcon, Mistral, and Llama-2 fine-tuning (some Llama-2 validation still remains). The fine-tuning wrapper is constructed with a focus on extensibility, allowing for rapid expansion to other LLMs assuming users hold a minimal level of LLM-specific knowledge. 

\medskip
\noindent\textbf{Multiple Prompt Structures}: 
Translation quality, as demonstrated by numerous prior works, varies significantly with prompt structure. As such, the fine-tuning wrapper of Simul-LLM supports multiple kinds of prompt structures, all of which we will validate later in this paper. In the interest of supporting adapting NMT LLMs to SimulMT, Simul-LLM supports NMT fine-tuning in addition to supporting prompts that allow for strict \textit{wait-k} fine-tuning structures. 

\begin{figure*}[ht]
    \centering
    \includegraphics[width=\textwidth]{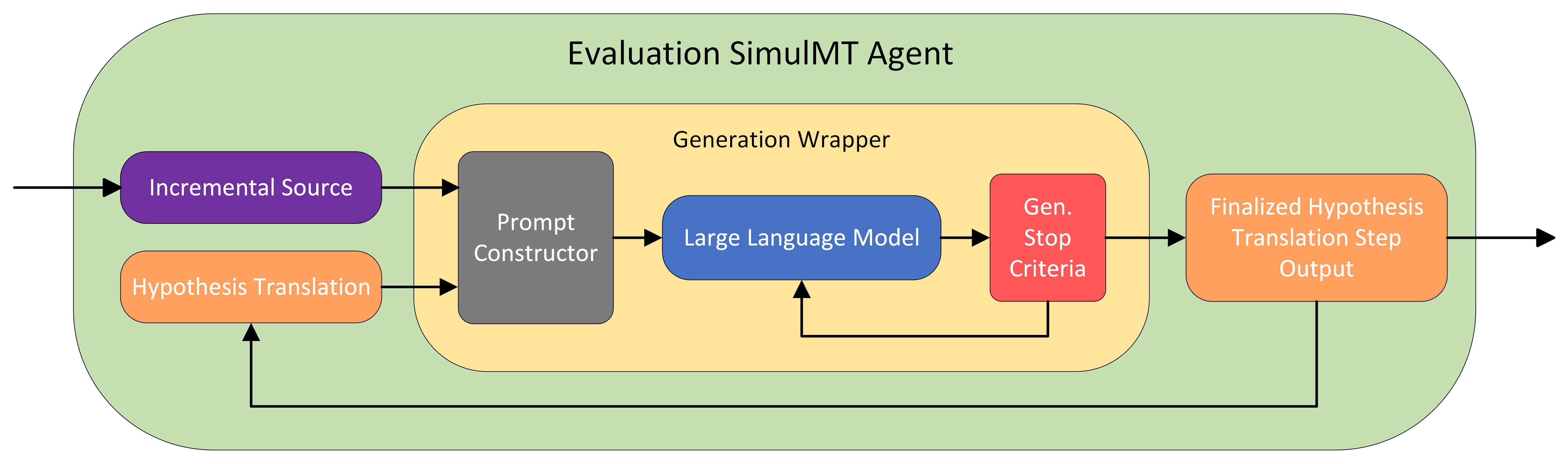}
    \caption{Depiction of the Simul-LLM evaluation agent framework. The SimulMT agent receives the incremental source from SimulEval (left of the figure) and sends the finalized translation step hypothesis to SimulEval (right of the figure), which manages lagging/latency calculation and translation quality scoring.}
    \label{fig:agent_and_stop_crit}
    \vspace{-1em}
\end{figure*}


\medskip
\noindent\textbf{Flexible Quantization and PEFT}: 
Effectively fine-tuning LLMs is reliant on careful memory management for low-to-mid hardware setups. Given that, Simul-LLM quantizes LLMs via the \verb|BitsAndBytes| library (seen in the Quant Config in Figure \ref{fig:finetune}), which enables flexible fixed-point quantization. For most low-to-mid hardware setups, we recommend quantizing in 4-bit floating point via 4-bit NormalFloat (nf4). Additionally, Simul-LLM by default engages in PEFT \cite{peft} via a passed configuration, although it does also support full model fine-tuning. 

\medskip
\noindent\textbf{Prompt Loss Filtering}: 
Extremely basic supervised fine-tuning may inappropriately include portions of the prompt in loss calculations that are then backpropagated. The fine-tuning wrapper of Simul-LLM ensures that the model only learns from data it is intended to generate post-prompt via a \verb|DataCollator| object and a response template. 


\subsection{Evaluation Agent and Features}
Evaluation agents for Simul-LLM are similarly built for ease of use and extensibility in addition to customizability towards complex translation schedules and decoding strategies while seamlessly interfacing with the preeminent SimulMT evaluation framework, SimulEval \citep{simuleval2020}. This is depicted in Figure \ref{fig:agent_and_stop_crit} and includes the following features.

\medskip
\noindent\textbf{Classical SimulMT Translation Scheduler}: 
In the interest of baseline accessibility, Simul-LLM evaluation agents support \textit{wait-k} translation schedules for SimulMT, given their ease of application. Further adaptive or otherwise more involved translation schedulers can be quickly constructed and applied, assuming no reliance on fine-tuning.

\medskip
\noindent\textbf{Support for Multiple Decoding Strategies}: 
Variable latency constraints for possible inference environments demand flexibility in decoding strategies. As such, evaluation agents for Simul-LLM support several decoding strategies, including greedy and naive decoding, subword-based beam search for single-word decoding, and variations on Speculative Beam Search (SBS) \citep{zheng-etal-2019-speculative}, including single word-based and chunk-wise SBS.


\medskip
\noindent\textbf{Scoring and Latency via SimulEval}: 
SimulEval \citep{simuleval2020} is the premier SimulMT evaluation framework. Simul-LLM evaluation agents interface seamlessly with SimulEval, which handles the incremental source context and manages translation scoring and latency tracking regarding the translation hypothesis. In addition to the core functionality of SimulEval in terms of translation quality and latency scoring, Simul-LLM also features some extensions to SimulEval. This includes the capability to measure computationally aware text-based latency via the specification of an oracle, zero-latency transcription agent (effectively required for latency modeling, elaborated on in our Appendix) and support for COMET \citep{rei-etal-2020-comet} translation quality evaluations (default model employed is COMET-DA \citep{rei-etal-2021-references}).



\section{Adapting NMT LLMs to SimulMT}
\label{sec:adapt}
Existing LLMs that have been fine-tuned for classical NMT may have the potential to be employed directly for SimulMT inference. This can be desirable under circumstances where a single deployed model is preferable to multiple, specialized models (e.g. avoiding fine-tuning costs for multiple models). However, exactly how to adapt such models is unclear in practice given the differences between prompts during NMT fine-tuning (full-sentence availability) and SimulMT inference (incremental source availability). This is especially problematic when engaging with short \textit{wait-k} schemes and similar low-lagging schedules, where minimal source context is available during early translation steps. 

To intuit why this is an issue, suppose that an NMT LLM is accustomed to receiving the entire source sequence $\textbf{x}$ before outputting a word or token $\textbf{y}_1$. If engaging in a low-lagging \textit{wait-k} where $|\textbf{x}| >> k$, then the output $\textbf{y}_1$ can now only be based on source context up to $\textbf{x}_k$. Assuming the NMT LLM would typically rely on some source context $\textbf{x}_i$ where $i > k$, then critical information is missing from its translation decision. We have conducted preliminary quantitative studies on this front by employing the proposed Simul-LLM framework, and these results are presented in Section \ref{sec:results}.

\section{Prompt Structure for SimulMT LLMs}
\label{sec:prompt}
Alternative to adapting NMT LLMs to SimulMT tasks, the SimulMT LLM approach aims to fine-tune LLMs directly for SimulMT, which requires new prompt structuring. 
Unlike classical encoder-decoder models for translation, source context for an LLM must be packaged within a prompt and structured appropriately. Some existing work has studied prompt structures for typical NMT \citep{zhang2023prompting, xu2023paradigm}, but it is unclear whether such prompts are still optimal for SimulMT. This is because significant differences exist in source context availability between fine-tuning and inference, as discussed in Section \ref{sec:adapt}. 

To address this need, we propose a new prompting approach to construct the dataset for fine-tuning and evaluation of SimulMT LLMs.
Instead of structuring each source-target sentence pair as a single example as typically done for NMT, we propose decomposing and expanding every sentence pair into a number of examples, where each example incrementally provides additional source and target tokens. When employed, the expanded examples directly mimic the behavior of simultaneous translation. We investigate two specific structures of materializing this prompt structuring approach, as analyzed and compared below.

\begin{figure*}[ht]
    \centering
    \includegraphics[width=\textwidth]{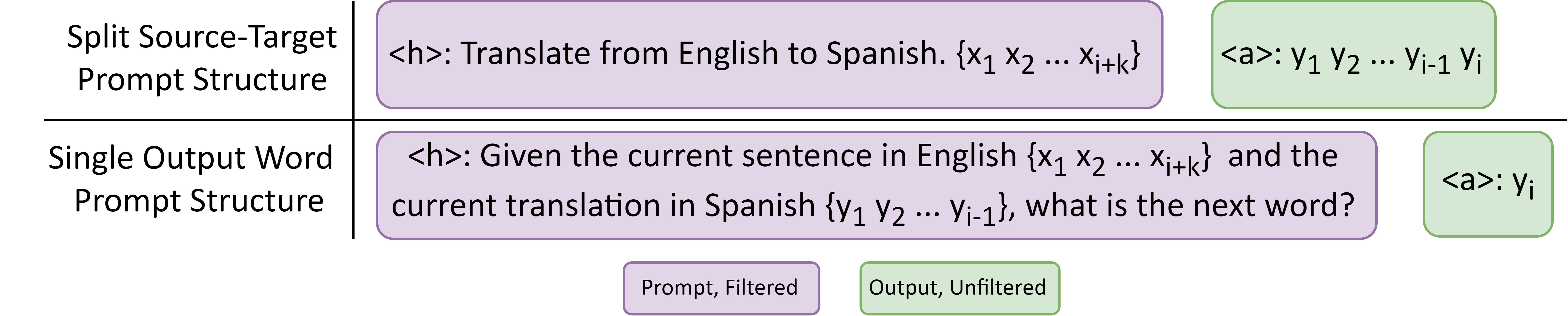}
    \caption{Example of English to Spanish translation prompt construction with an incremental source \textbf{x} and an incremental output \textbf{y} applied via our proposed expanded dataset. Without more complex loss filtering than is typical, the entire output sequence for the split source-target prompt structure would be scored and the model would learn for \textit{wait-k} schedules ranging from \textit{wait-i} to \textit{wait-k} as opposed to just \textit{wait-k}.}
    \label{fig:prompt_diagram}
    \vspace{-1em}
\end{figure*}



\subsection{Split Source-Target Prompt Structure}
The first prompt structure follows the classical NMT prompts where the source sentence is included in the prompt and the target sentence is the output of the LLM. As illustrated in the first prompt structure in Figure \ref{fig:prompt_diagram}, with the expanded examples, each example contains partial source along with some instruction in the prompt (starting with "<h>") and partial target that is \textit{k} words behind the source in the model output (starting with "<a>"), where \textit{k} is the intended inference \textit{wait-k} value. While this \textit{split source-target} prompt structure seems to be a natural and plausible way for model fine-tuning, it has exhibited difficulties when learning for simultaneous translation. 
The root of this stems from how fine-tuning with LLMs often works: when filtering the prompt from loss calculations during fine-tuning, a response template of some kind (e.g. "<a>:" or ">>ANSWER<<") is employed to ensure only the target word that is generated at the current step (i.e., $\textbf{y}_i$) is scored. Unfortunately, with the split source-target prompt,
the template allows for all the target words that have been generated from previous translation steps (i.e., $\textbf{y}_1$ to $\textbf{y}_i$) to be scored. Without employing a more complex loss filtering, this leads to an inappropriate level of context.

This lack of loss filtering can be especially problematic near the end of a given sequence's translation. 
As a simple example, suppose that a given source sequence is of length $|\textbf{x}|$ and $|\textbf{x}| >> k$. In the first prompt structure where up to $i$ source words have been supplied and where $|\textbf{x}| > i >> k$, the LLM is effectively being fine-tuned for varying \textit{wait-k} values ranging from \textit{wait-i} to \textit{wait-k} in a single example (as $\textbf{y}_i$ is predicted from $\textbf{x}_1$ to $\textbf{x}_{i+k}$ which is \textit{wait-k}, $\textbf{y}_{i-1}$ is predicted from $\textbf{x}_1$ to $\textbf{x}_{i+k}$ which is \textit{wait-$(k+1)$}, and so on). If $i >> k$ to the point where it is close to normal NMT levels of context, where $i \: \: \Tilde{=} \: \: |\textbf{x}|$, then the LLM is no longer being effectively fine-tuned with an appropriate amount of source context for a given translation (write) decision schedule. 


\subsection{Single Output Word Prompt Structure}
\label{sec:single_out}
The above problem can be entirely side-stepped by our proposed second approach, \textit{single output token} prompt structure, that embeds only the current target translation hypothesis within the model output. As illustrated in the second prompt structure in Figure \ref{fig:prompt_diagram}, instead of allowing target translation hypotheses from previous time-steps to be incorporated into the loss, the proposed prompt structure shifts those previous translation hypotheses into the prompt. Combined with the expanded examples that form a rigorous \textit{wait-k} curriculum in terms of the fine-tuning dataset (rigorous meaning a complete curriculum as opposed to a random subset), inference behavior can be copied exactly for every fine-tuning example, completely closing the context mismatch between fine-tuning and inference. 

Between the two prompt structures, it is clear that the single output word prompt structure more closely replicates the relationship observed in Equation \ref{eq:waitk} between the source and target sequence during fine-tuning. For both structures, a source-target sentence pair is expanded up to $max(\textbf{|x|} - (k-1), \textbf{|y|})$ examples.

\section{Evaluation Methodology}
\label{sec:methods}
To validate our proposed solutions to the aforementioned challenges and to test the capabilities of Simul-LLM as the first SimulMT LLM open-source framework, we engaged in several experiments allowing for comparisons among classical non-LLM-based NMT \citep{attention_is_all_you_need} / SimulMT architectures \citep{stacl2019}, NMT LLMs adapted for SimulMT, and SimulMT LLMs. All mentioned LLMs are fine-tuned Falcon-7B models, but Simul-LLM features an easy to extend framework and we support both Llama and Mistral-based models as well (results in Appendix). 

NMT LLMs were fine-tuned for one epoch, as overfitting was observed beyond that point, which we intuit to be possibly due to the well-documented ability of LLMs to quickly memorize training sets \citep{biderman2023emergent}. In contrast, SimulMT LLMs were fine-tuned for either 2M random examples out of 5M examples (for \textit{wait-3} fine-tuning) or up to 5M examples (for \textit{wait-7} fine-tuning) on the expanded dataset due to computational constraints.


\begin{table*}[t]
    \centering
    \begin{tabular}{l|l|cc}
        \hline
         \multirow{2}{2.5cm}{Grouped Explorations} & \multirow{2}{*}{Model and Decoding Scheme} & \multirow{2}{*}{en-de} & \multirow{2}{*}{en-es} \\ & \\
        \hline
        \multirow{2}{2.5cm}{Classical Baselines} & NMT Transformer (non-simultaneous) & 26.96 (22.6) & 32.64 (23.1) \\
        & Monotonic Transformer \textit{Wait-5} (SimulMT) & 22.01 (3.32) & 24.90 (2.58) \\
        \hline
        \multirow{6}{2.5cm}{NMT LLMs Adapted for SimulMT} & NMT LLM & 25.83 (3.65) & 30.06 (3.95) \\
        & NMT LLM Single SBS (k=3, b=5, c=1, w=6) & \underline{25.98 (4.12)} & 29.48 (4.64) \\
        & NMT LLM Single SBS (k=3, b=5, c=1, w=10) & 25.95 (4.25) & 27.67 (4.82) \\
        & NMT LLM Chunk SBS (k=3, b=5, c=2, w=10) & 23.61 (4.63) & 26.33 (5.26) \\
        & NMT LLM Chunk SBS (k=5, b=5, c=3, w=15) & 25.80 (5.61) & 27.00 (5.90) \\
        & NMT LLM Chunk SBS (k=7, b=5, c=4, w=20) & \textbf{27.32 (6.97)} & 28.66 (7.09) \\
        \hline
        \multirow{3}{2.5cm}{SimulMT LLMs with Proposed Prompt} & \textit{Wait-3} Fine-tuning LLM (2M samples) & 19.99 (3.41) & 23.68 (3.64) \\
        & \textit{Wait-7} Fine-tuning LLM (2M samples) & 20.82 (3.44) & 25.18 (3.61) \\
        & \textit{Wait-7} Fine-tuning LLM (5M samples) & 21.86 (3.38) & \underline{30.31 (3.34)} \\
        & \textit{Wait-7} Fine-tuning LLM (2M samples, k=7) &  23.09 (6.71) & 28.92 (6.87) \\
        & \textit{Wait-7} Fine-tuning LLM (5M samples, k=7) &  24.93 (6.70) & \textbf{35.14 (6.73)} \\
    \end{tabular}
    \caption{Comparisons of performance for various models and decoding schemes during primarily \textit{wait-3} evaluation (non-\textit{wait-3} is specified via k) via detokenized BLEU. 
    Non-LLM baselines are subword-based \textit{wait-k} (standard) while LLMs are word-based \textit{wait-k}. Best SimulMT quality results are \textbf{bolded}, second best results are \underline{underlined}, and lagging values are provided in parentheses as LAAL \citep{Papi_2022}. Speculative Beam Search (SBS) during inference is experimented with for NMT LLMs, which lend themselves towards SBS (k=\textit{wait-k} value, b=beams, c=chunks/words, w=window size). 
    }
    \label{tab:comparisons}
\end{table*}

\subsection{Dataset Selection and Preprocessing}
No standardized dataset exists for SimulMT with LLMs. Due to its popularity in speech-to-text simultaneous translation (SimulST), we employ MuST-C for our experiments\footnote{This allows for future work that explores multi-modal simultaneous LLMs, engaging in SimulST via a cascaded model structure with a transcription model for the source speech or a joint speech/text-to-text framework.}. For the purposes of adapting MuST-C for text-to-text usage, we preprocess the dataset and filter out certain acoustic indicators (e.g., floating "\textsf{-}" characters representing pauses). In some cases, this resulted in significant changes to some samples of the test set, such as the removal of \small \textsf{(Applause)} \normalsize acoustic indicators. 


We employ MuST-C across two language pairs, those being English-to-German (en-de) and English-to-Spanish (en-es). Some additional experiments are provided for the en-es language pair that validate fundamental SimulMT concepts and display BLEU scores, gathered via sacreBLEU \citep{sacrebleu2018}, with respect to samples observed during fine-tuning. The original dataset contains roughly 270K training set samples and approximately 2.5-3K test set samples (tst-COMMON split) per language pair. The expanded version of this dataset for single output word fine-tuning contains approximately 5M training set samples per language pair.

\subsection{Word or Token-Based \textit{Wait-k} for LLMs}
While classical encoder-decoder SimulMT systems usually engage in either word or token-based \textit{wait-k}, they most typically engage with whichever is more suitable for their vocabulary (i.e. word versus sub-word vocabularies). In spite of the fact that LLMs function via sub-word vocabularies, we recommend, and employ for this work, word-based \textit{wait-k} for SimulMT LLMs, as it more closely resembles the flow of engaging with a natural language interface. Moreover, supposing that the LLM is receiving a given sequence actively from a transcription system or something similar, it makes intuitive sense to wait for a word to be emitted from the system as opposed to a fragment.  

\section{Results and Analysis}
\label{sec:results}
\subsection{Exploration of Adapting NMT LLMs to SimulMT}
In Table \ref{tab:comparisons}, we provide a breakdown of the performance of several different models, decoding strategies, and \textit{wait-k} schedules. Regarding our exploration related to adapting NMT LLMs to SimulMT, we also include results related to our implementation of Speculative Beam Search (SBS) \citep{zheng-etal-2019-speculative}. As demonstrated by these results, compared with classical models, LLMs fine-tuned for NMT are very capable of SimulMT upon being adapted during inference (even exceeding the score of the classical NMT transformer on en-de that performs non-simultaneous translation). It is worth reiterating that classical architectures often engage in subword-based \textit{wait-k} whereas we employ word-based \textit{wait-k} for LLMs, but the comparisons still serve as a useful reference.

SBS-based decoding strategies helped NMT LLMs in the en-de language pair, but lacked improvement for the en-es language pair. We noted that our implementation (and seemingly also the original implementation) was sensitive to both the window size and the number of committed chunks, with large values for either resulting in the speculative target translation getting too close to the size of the source context. In our tests, when reaching the same length as the source context (akin to context levels of \textit{wait-1}), degenerate output began to appear that resulted in the output trailing off (e.g., final output of "que..." instead of correct output of "que"). Notably, too many committed chunks only explains performance gaps for chunk-wise SBS, not single SBS, which is normally a flat improvement upon greedy decoding (single SBS is still sensitive to window size). Future experiments can be conducted to utilize the proposed Simul-LLM framework to quantify these factors.

\subsection{Exploration of SimulMT LLMs with Proposed Prompt Structure}
In Table \ref{tab:comparisons}, we also provide a breakdown of the performance of our exploration of SimulMT LLMs with our proposed prompt structure in Section \ref{sec:single_out} that carefully manages source context availability for target translation generation. Two models are employed for this exploration, one fine-tuned for \textit{wait-3} inference, and another fine-tuned for \textit{wait-7} inference that produces noticeably better quality translations than the first. In addition, we provide results for the fine-tuned \textit{wait-7} models after fine-tuning on 2M samples and 5M samples of the fine-tuning dataset, with the fine-tuned \textit{wait-3} model having fine-tuned for 2M samples. This separation is due to computational constraints on our side, as we wanted to provide further comparisons in Section \ref{sec:waitk_gen} and Table \ref{tab:ft_to_inf} at various evaluation \textit{wait-k} values but lacked the compute to finish fine-tuning the \textit{wait-3} model.

While SimulMT LLMs are a more promising approach in achieving higher translation quality than NMT LLMs due to more direct task-specific fine-tuning and better context alignment, our experimental results suggest that, for the time-being, the performance of SimulMT LLMs is not guaranteed to be advantageous compared to NMT LLMs adapted to SimulMT. As seen in Table \ref{tab:comparisons}, fully fine-tuned SimulMT LLMs outperformed their NMT LLM counterparts on en-es but failed to do so on en-de. This is not completely unexpected as NMT LLMs have been optimized heavily in recent years whereas the exploration of SimulMT LLMs has just started. We provide some analysis below that points out a few possible reasons for this observed performance gap and we call for additional community efforts to investigate further.

First, it is possible that the fine-tuning hyperparameters are ill-suited for this particular prompt. We consider this likely to be the most influential issue on our observed results, given the drastic differences between the original and expanded datasets (the fine-tuning hyperparameters were united for both fine-tuning tasks). Second, at least one other work related to NMT LLMs \citep{chen2023improving} has demonstrated that relative positional embeddings can cause issues via attention dilution that ends up being unhelpful, suggesting that distancing the source context, running target hypothesis, and the current translation step hypothesis can be unexpectedly problematic. We posit that our proposed Simul-LLM can be leveraged to verify the above reasons.

\begin{table}[t]
    \centering
    \begin{tabular}{l|l|c}
        \hline
        \multirow{2}{2cm}{Fine-tuning \textit{Wait-k}} & \multirow{2}{1.5cm}{Inference \textit{Wait-k}} & \multirow{2}{1cm}{BLEU} \\&\\
        \hline
        \multirow{3}{2.5cm}{SimulMT LLMs Fine-tuned in \textit{Wait-3}} & \textit{Wait-3} & 23.68 \\
        & \textit{Wait-5} & 25.59 \\
        & \textit{Wait-7} & 26.31 \\
        \hline
        \multirow{3}{2.5cm}{SimulMT LLMs Fine-tuned in \textit{Wait-7}} & \textit{Wait-3} & 25.18 \\
        & \textit{Wait-5} & 28.19 \\
        & \textit{Wait-7} & 28.92 \\
        
    \end{tabular}
    \caption{BLEU scores for various SimulMT LLMs fine-tuned with different \textit{wait-k} values on en-es. All models fine-tuned for 2M samples due to computational constraints.}
    \label{tab:ft_to_inf}
\end{table}

\subsection{Higher \textit{Wait-k} Generalizability Comparisons}
\label{sec:waitk_gen}
It is well documented that in typical SimulMT systems, training or fine-tuning with a slightly higher \textit{wait-k} than intended during inference can boost translation quality and generalizability across slightly lower \textit{wait-k} \citep{stacl2019}. While this likely applies to SimulMT LLMs, no existing work has validated that this behavior persists. We provide a brief comparison of two SimulMT LLMs fine-tuned via \textit{wait-3} and \textit{wait-7} context levels in Table \ref{tab:ft_to_inf}. The results demonstrate that, generally, the expected behavior does hold, with all LLMs fine-tuned in \textit{wait-7} outperforming their corresponding \textit{wait-3} models for the same inference \textit{wait-k} with up to a 2.6 BLEU improvement. We leave validating additional, previously understood SimulMT principles in SimulMT LLMs to future work.

\subsection{Discussion on Other Works}
We are aware of two other concurrent efforts at applying LLMs to simultaneous translation \citep{koshkin2024transllama, wang2023simultaneous}. The former focuses on high-resource settings (i.e. order-of-magnitude larger language models than we aim for in this work) and additionally touches on comparisons with GPT-4 and speech-to-text simultaneous translation efforts. Contrastingly, the latter work aligns with ours, approaching the problem of LLM simultaneous translation in a low-resource setting. Their work, however, is limited to a focused contribution: a variation of longest-common prefix (LCP) that they call relaxed agreement LCP (RALCP). Neither host open-source software to test their changes and both only test Llama-based models in addition to arguably shallower explorations of this subject matter. Nonetheless, we intend to add support for end-to-end speech-to-text translation soon and have already added support for RALCP, which we provide brief results for in our Appendix. Both works validate our choice to engage with MuST-C as a dataset, as both works also employ it.

\section{Conclusion}
In this work, we introduce Simul-LLM, the first open-source framework that enables rapid development of LLM fine-tuning and evaluation pipelines for simultaneous machine translation (SimulMT). Simul-LLM seamlessly integrates with the fine-tuning and generation tools of the popular \verb|transformers| library as well as with SimulEval, the preeminent SimulMT evaluation framework. In addition to introducing Simul-LLM, we employ this framework to explore a wide range of topics in the LLM SimulMT space. Our proposed Simul-LLM framework enables multiple lines of future work that can be carried out to understand and optimize LLMs for simultaneous translation, and it will likely be a useful tool for the research community.


\section{Limitations}
This work is focused on providing an open-source framework for an interesting application of LLMs. While we propose a few novel techniques for engaging with LLMs in this application space that we support in Simul-LLM, the work is largely exploratory and we cannot assert that our proposed practices are guaranteed to work best. Additionally, the framework itself does not currently support speech-to-text exploration (although augmentation to this end should not be difficult), a popular simultaneous translation track. All of these points serve to slightly limit the impact of our contribution.

\section{Acknowledgements}
This research was supported, in part, by the National Science Foundation grants 2223483 and 2223484.

\bibliography{anthology,custom}

\appendix

\section{Appendix}
\label{sec:appendix}

\subsection{Licensing Information}
Fairseq \citep{fairseq2019} is MIT-licensed. SimulEval is licensed via CC BY-SA 4.0. MuST-C, one of the premiere speech-to-text datasets across many language pairs, is licensed under CC BY-NC-ND 4.0. Simul-LLM itself is MIT-licensed.

\subsection{Training, Fine-Tuning, and Evaluation Hyperparameters and Hardware Details}
All classical models were trained on two NVIDIA 32GB V100s and validated on a single V100. All LLMs were fine-tuned via PEFT \cite{peft} on a single NVIDIA 40GB A40 in bfloat16 and evaluated on a single V100 in float32. Simul-LLM seamlessly integrates with SimulEval \citep{simuleval2020} for the purpose of these evaluations. Classical transformer baselines were trained via Fairseq \citep{fairseq2019}, an easily extensible sequence to sequence modeling toolkit written in PyTorch.

Classical models were trained with typical hyperparameters provided in Fairseq examples. All LLMs were fine-tuned with identical hyperparameters, employing a constant learning rate of 3e-4 and were optimized via Adam with around 4K warmup updates and batch sizes of 40 samples. LoRA adapter parameters were an $\alpha$ of 16 and a $r$ value of 64, resulting in a total of around 40M added parameters during fine-tuning, with a dropout value of 0.1. For fair comparison, classical models were of a similar, although not quite identical, size in terms of parameter count (around 46M parameters). 

Additionally, all LLMs were fine-tuned while quantized with NormalFloat4 (nf4) quantization. A small performance boost was observed when removing this quantization during inference, so all models did not engage with nf4 quantization during inference.

\subsection{Estimated Computational Costs}
We estimate that, normalizing for 1 GPU, building Simul-LLM cost roughly 26 GPU days of experimentation and approximately 24 GPU days to generate the results in this paper. 

\begin{table}[t]
    \centering
    \begin{tabular}{l|c}
        \hline
        Model & CA LAAL (ms) \\
        \hline
        Monotonic Transformer \textit{Wait-5} & 1325 \\
        \textit{Wait-7} Fine-tuning LLM & 3237 \\
    \end{tabular}
    \caption{Brief results for modified computationally aware Length-Adaptive Average Lagging (in milliseconds). Inherent, estimated acoustic latency is set to roughly 360 milliseconds.}
    \label{tab:ca_laal}
\end{table}

\subsection{Computational Latency of LLMs for SimulMT}
While we only cover results for lagging behavior (via LAAL) in the main body of this paper, it is worth acknowledging that employing an LLM for SimulMT incurs additional computational latency costs compared to a more classical solution. These costs can be characterized in a number of ways, but we choose to employ a latency-based version of LAAL inspired by simultaneous speech translation (SimulST), where we replace word-based lagging measurements with estimated acoustic latency in milliseconds. This allows us to simply add the computational cost of each generation step to the inherent, estimated acoustic latency (we employ 360 ms of inherent latency, equivalent to around 166 words/min). Adding this inherent latency is akin to treating the SimulMT system as though an oracle, zero-latency transcription agent has been prepended to it and is critical to ensuring that LAAL operates correctly, as a metric. Brief latency results can be observed in Table \ref{tab:ca_laal}. For further details, we refer readers to the original LAAL publication \citep{Papi_2022}.

\begin{table}[t]
    \centering
    \begin{tabular}{l|cc}
        \hline
        Decoding Scheme & en-de & en-es \\
        \hline
        Chunk SBS & 25.8 (5.6) & 27.0 (5.9) \\
        Chunk SBS w/ RALCP & 26.1 (5.3) & 28.1 (5.5) \\
    \end{tabular}
    \caption{Demonstration of RALCP support and effectiveness via BLEU and LAAL (provided in parentheses) with k=5, b=5, c=3, and w=15 as per Table \ref{tab:comparisons}. Agreement factor is set to 0.6. All models here employ the NMT LLM prompt structure mentioned in this paper.}
    \vspace{-1em}
    \label{tab:ralcp}
\end{table}

\subsection{RALCP and Policy Differences}
While Simul-LLM supports RALCP \citep{wang2023simultaneous}, our application differs slightly from the specification of the original implementation. Unlike the authors of \citealt{wang2023simultaneous}, we do not engage in consecutive read decisions unless it is to maintain a \textit{k}-lagging factor as specified by an inference \textit{wait-k} value. This ensures, after a given set of outputted chunks via RALCP implemented on top of SBS \citep{zheng2020simultaneous}, that expected context limitations are kept. This contrasts with the original implementation of RALCP on top of a more relaxed read-write schedule that does not always attempt to maintain a \textit{k}-lagging factor after that lagging factor is initially established. Additionally, we assume that the authors applied a word-based simultaneous approach, as this is not specified in their original work.

Nonetheless, RALCP, which is fundamentally a form of beam rescoring, does improve translation quality based on our empirical results. We include a brief comparison in Table \ref{tab:ralcp}, where we observe that the addition of RALCP seems to result in small to moderate improvements in BLEU score with roughly equivalent lagging behavior (slight reductions in our tests).

\subsection{Supporting Various LLMs}
Simul-LLM's easy to extend fine-tuning and evaluation framework supports Falcon, Llama, and Mistral-based models. While our explorations in this paper are rooted in Falcon-based models, we provide a brief set of results for other LLMs in Table \ref{tab:llm_comp}. Generally, translation quality appeared similar across models, although the Llama-based model struggled. BLEU degradation for Llama models seemed mostly linked to degenerate output where the model would get stuck in low-quality generation (e.g. continuously outputting the same token).

\begin{table}[t]
    \centering
    \begin{tabular}{l|c}
        \hline
        Foundational Model & BLEU (LAAL) \\
        \hline
        Falcon 7B LLM & 30.06 (3.95)\hphantom{0} \\
        Mistral 7B LLM & 30.25 (3.66)\hphantom{0} \\
        Llama-2 7B LLM & 20.18 (3.84)\tablefootnote{Some issues were observed during fine-tuning the Llama-based translation agent that seems to have resulted in poorer than expected translation quality. We are actively working to improve these results within the Simul-LLM framework.} \\
    \end{tabular}
    \caption{Brief comparison of Falcon, Mistral, and Llama-based models fine-tuned and evaluated in Simul-LLM on en-es. All models were fine-tuned with the NMT LLM prompt structure specified in this paper and are evaluated via a \textit{wait-3} schedule with greedy decoding on en-es.}
    \vspace{-1em}
    \label{tab:llm_comp}
\end{table}

\end{document}